# An Evolutionary approach for solving Shrödinger Equation

Khalid jebari[1], Mohammed Madiafi[2] and Abdelaziz Elmoujahid[3]

[1] LCS Laboratory Faculty of Sciences Rabat Agdal, University Mohammed V, UM5A
Rabat, Morocco
*khalid.jebari@gmail.com*

[2] Faculty of Sciences Ben M'Sik, University Hassan II Mohammadia, UH2M
Casablanca, Morocco
*Madiafi.med@gmail.com*

[3] LCS Laboratory Faculty of Sciences Rabat Agdal, University Mohammed V, UM5A
Rabat, Morocco
*elmoujahid@gmail.com*

**Abstract**

The purpose of this paper is to present a method of solving the Shrödinger Equation (SE) by Genetic Algorithms and Grammatical Evolution. The method forms generations of trial solutions expressed in an analytical form**.** We illustrate the effectiveness of this method providing, for example, the results of its application to a quantum system minimal energy, and we compare these results with those produced by traditional analytical methods.

## 1. Introduction

One of the fundamental equations in modern physics is the Schrodinger equation [4]. This equation is used to model quantum mechanical systems that may occupy several different states which correspond to different energy levels, and on which all we can hope to know is the probability of presence in each possible state at every moment. The Schrödinger equation is usually written as follows:

$$H \times \Psi(r,t) = E \times \Psi(r,t) \qquad (1)$$

Where H is the Hamiltonian operator of the system studied, E its total energy and $\Psi(r, t)$ the wave function that depends on space coordinates, r and time t. The Hamiltonian is usually written as the form:

$$H = \left( \frac{-\hbar}{2m} \nabla^2 + V(r,t) \right) \qquad (2)$$

Where $\hbar$ denotes Planck's constant, m is the mass of the system, $\nabla$ the gradient operator and V potential of the system. It is therefore a partial differential equation that describes the evolution in time of wave function of a physical system, but we cannot be solved analytically, except in very simple special cases [6]. Faced to this difficulty of obtaining accurate analytical solutions, physicists are content generally approximate solutions, but that may well be satisfactory, using numerical techniques.

In this paper we propose and discuss a different approach to approximate solution of the Schrödinger Equation. It is an approach which is based on genetic algorithms [5] whose principle is briefly described at the beginning of Section 2. Our method offers analytical form solutions; however the variety of the basis functions involved is not a priori determined, rather is constructed dynamically as the solution procedure proceeds and evolve. The generation of solutions progress with using a grammatical evolution (GE), governed by a grammar expressed in Backus Naur Form (BNF) [10]. GE described in section 3. The Section 4 will be devoted to the presentation of the proposed method. The numerical results and discussion will be the subject of Section 5 Finally Section 6 contains a conclusion to this work.

## 2. Genetic Algorithms

Genetic algorithms (GA) are search methods and stochastic optimization inspired by Darwin's theory on the evolution of species [5]. These methods are theoretically applicable to any complex problem can be posed mathematically as a research problem of a point, which represents the best solution possible in a search space, which represents the set of all candidate solutions. In practice, however, in order to develop a genetic solution to a problem we must first find a way to encode mathematical solutions of this problem as data structures called chromosomes [5]. Then a fitness function, often called fitness, is necessary to measure and compare the relative quality of different solutions candidates. Thus, starting from an initial population of solutions, which can be chosen randomly, a AG allows, through iterations called generations, to change the individuals in this population towards solutions increasingly better, and this in accordance with rules selection, reproduction and mutation that mimic biological evolution. More formal, a GA can be described simply using the following pseudo code [9]:

```
/ / GA algorithm
• Choose an initial population of individuals: P (0)
• Assess the degree of adaptation of each individual of P (0)
• Set a maximum number of generations not to exceed: max t
• While (we are not satisfied and t max <t) {Making
- T ← t +1
- Select parents crossed to produce children
- Apply the operators of crossover and mutation
- Create a new population of survivors: P (t)
- Evaluate P (t)
}
 Return the best individual of  P(t)
```

Fig. 1 Pseudo code of Standard Genetic Algorithms

GAs has proven their effectiveness in solving many complicated problems of the world real. These problems cover several application areas including bioinformatics, industry, medicine, economics, robotics, etc.. And in this work we propose to use Gas to solve the Schrödinger equation in accordance with the procedure described in the proposed method section.

## 3. Grammatical Evolution

Grammatical evolution (GE) is an evolutionary algorithm that can produce code in any programming language. GE has been applied successfully to problems such as symbolic regression [11], discovery of trigonometric identities [12], robot control [13], caching algorithms [14], and financial prediction [11].
 The algorithm requires as inputs the BNF grammar definition of the target language and the appropriate fitness function. Chromosomes in grammatical evolution, in contrast to classical genetic algorithms, are expressed as vectors of integers. Each integer denotes a production rule from the BNF grammar. The algorithm starts from the start symbol of the grammar and gradually creates the string, by replacing non terminal symbols with the right hand of the selected production rule. The selection is performed in two steps:

[0]   We read an element from the chromosome

(with value V ).

[1]   We select the rule P according to the scheme

$$P = V \bmod R \quad (4)$$

Where R is the number of rules for the specific non terminal symbol.

A BNF grammar is made up of the tuple N, T, P, S; where N is the set of all non-terminal symbols, T is the set of terminals, P is the set of production rules that map N to T, and S is the initial start symbol and a member of N. Where there are a number of production rules that can be applied to a non-terminal, a " | " (or) symbol separates the options. Using:

N = {<expr>, <op>, <operand>, <var>, <func>}
T = {1, 2, 3, 4, +, -, /, *, x, y, z}
S = {<expr>}

The process of replacing non terminal symbols with the right hand of production rules is continued until the end of chromosome has been reached. We can reject the entire chromosome or we can start over (wrapping event) from the first element of the chromosome if threshold of the number of wrapping events is reached, the chromosome is rejected by assigning to it a large fitness value. In our approach, threshold of the number of wrapping events is equal to 2.

As well we programmed in C++ language, we used a small part of the C programming language grammar as we can see an example in Table II, we also construct 4 Radial Basis Functions (RBF), see Table I, that we added to the C++ standard functions. We note that the numbers in parentheses denote the sequence number of the corresponding production. Below is a BNF grammar used in our method:

```
        S::=<expr>            (0)
        <var>::=x             (0)
           |y          (1)
           |z          (2)
        <operand> ::= 0       (0)
           | 1        (1)
           | 2        (2)
           | 3        (3)
           | 4        (4)
           | 5        (5)
           | 6        (6)
           | 7        (7)
           | 8        (8)
           | 9        (9)
           | <var>    (10)
```

```
<op> ::= +          (0)
       | -          (1)
       | *          (2)
       | /          (3)

<func> ::= sin      (0)
         |cos       (1)
         |exp       (2)
         |log       (3)
         |sqrt      (4)
         |BRF1      (5)
         |BRF2      (6)
         |BRF3      (7)
         |BRF4      (8)

<expr> ::= <expr> <op> <expr>   (0)
         | ( <expr> )           (1)
         | <func> ( <expr> )    (2)
         | <operand>            (3)
```

Table 1: Radial Basis Function

| Name Used | Function Name | Definition |
|---|---|---|
| BRF1 | Gaussian | $\exp(-cr^2)$ |
| BRF2 | Hardy Multiquadratic | $\sqrt{c^2+r^2}$ |
| BRF3 | Inverse Multiquadratic | $\sqrt{\dfrac{1}{c^2+r^2}}$ |
| BRF4 | Inverse Quadratic | $\dfrac{1}{c^2+r^2}$ |

Consider the chromosome C, the steps of the mapping procedure are listed in Table 2. The result of these steps is the expression sqrt(3/x).

Table 1. Radial Basic Function

| String_BNF | Chromosome | Operation |
|---|---|---|
| <expr> | 10,4,8,15,3,7,19,21,9 | 10 mod 4=2 |
| <func>(<expr>) | 4,8,15,3,7,19,21,9 | 4 mod 9=4 |
| sqrt(<expr>) | 8,15,3,7,19,21,9 | 8 mod 4=0 |
| sqrt(<expr><op><expr>) | 15,3,7,19,21,9 | 15 mod 4=3 |
| sqrt(<operand><op><expr>) | 3,7,19,21,9 | 3 mod 11=3 |
| sqrt(3<op><expr>) | 7,19,21,9 | 7 mod 4=3 |
| sqrt(3/<expr>) | 19,21,9 | 19 mod 4=3 |
| sqrt(3/<operand>) | 21,9 | 21 mod 11=10 |
| sqrt(3/<var>) | 9 | 9 mod 3=0 |
| sqrt(3/x) | | |

## 3. Proposed Method

To simplify the ideas, without loss of generalization, let us in the particular case of a mono-dimensional system which we want to minimize the energy quantum independently of time. The Hamiltonian of this system is:

$$H = \left(\frac{-\hbar}{2m}\frac{d^2}{dx^2} + V(x)\right)\Psi(x) \quad (5)$$

Where $x \in [a,b]$

The boundary condition $\int_a^b \Psi(x)dx = 1$ (6)

Solving the SE for this system is to find the wave function ψ (x) minimizes its energy quantum. The problem arises naturally as a problem research is to find, in the space of all possible functions, the function minimizes the energy of the system. To encode the solutions it suffices to use as chromosomes, the wave functions themselves, and to measure the relative quality of several candidate solutions simply use as function:

$$L(\Psi) = H\Psi - E\Psi \quad (5)$$

In order to conceive a genetic solution to PE we have to determine the encoding method. Then the fitness function is used for assessing and comparing the solutions candidates for PE. Thus, starting from an initial population of randomly generated individuals, we evolved this population toward better solutions according to the rules of selection strategy, crossover and mutation. The details are as follows

**Encoding Method:** the ordinal encoding scheme was used in the proposed method. Under this scheme, a serial number is assigned to each gene from 0 to s where s=50.

**Initial population size**: generally, the initial population size can be determined according to the complexity of the solved problem. A larger population size will reduce the search speed of the GA, but it will increase the probability of finding a high quality solution. The initialization of the population of chromosomes is generated by a random process. Each chromosome is represented as permutation $j_1, j_2,..., j_s$ of 1, 2,.., s. The process does not yield any illegal chromosome, i.e. the alleles of each gene of the chromosome are different from each others.

**Fitness function**: the fitness function is a performance index that it is applied to judge the quality of the generated

solution of SE. The steps for evaluating chromosome $C_i$ by fitness function are:

1. Choose T equidistant points in [a,b]
2. Construct a solution $S_i$ from $C_i$ by using grammatical evolution described earlier;
3. Calculate $f(S_i) = \sum_{j=1}^{T} L(C_i)$

**Selection operator:** in the selection operation, the chromosome with the lower fitness function value will have a higher probability to reproduce the next generation. The aim of this operation is to choose a good chromosome to achieve the goal of gene evolution. The most commonly used method is Tournament Selection. In this study, we used a modified Tournament Selection, which guards in each iteration the best individual. The following pseudo code summarizes the Selection Method.

```
// N: population size
T_alea: array of integer containing the indices of individuals
in the population
T_ind_Winner : an array of individuals indices 's who will be
selected
L_sorted : a list of all individual indices sorted in decreasing
fitness values
l = 0
k=0
For (i=0; i<k; i++)
{
        Shuffle T_alea ;
        For (j=0; j<N; j=j+k+1)
        {
                C1 = T_alea(j);
                For (m=1; m<k; m++)
                {
                        C2=T_alea(j+m);
                        if f(C1)< f(C2) C1 = C2
                }
        // f(Ci): Fitness of individual Ci
                T_ind_Winner(l) = C1
                T_ind_Winner(l+1) = L_sorted (k)
                l=l+2;
                k=k+1
        }
}
```

Fig.2 Tournament Selection Modified

**Crossover operation**: the goal is to combine two parent for generating better offspring, by probability decision.

In our study, a novel Homologous crossover [13] operator is used, which is inspired by molecular biology and proceeds as follows:

1. Each parent's gene is read from the left with sequential manner. If the rules selected are identical this region is labelled " region of similarity"
2. In the limits of the "region of similarity", the first two crossover points are selected. These points selected are the same on both parents
3. In other region i.e. " region of dissimilarity ", the crossover operator chooses the two second crossover points by executing the following rules:
4. The two second crossover points are then selected from the regions of dissimilarity by respecting the following steps:

   3. 1 in the " region of dissimilarity", a random crossover point is selected in the first parent;
   
   3. 2 on the second parent, the crossover point is selected if the gene is associated with the same type of non-terminal as the gene following the crossover point on the first parent;
   
   3. 3 this gene is located by generating a random crossover point. The process searches incrementally, until an appropriate point is found;
   
   3. 4 once the second crossover point has been determined, the process applies the simple two points crossover;
   
   3. 5 if no point is found in the second parent then the crossover operation did not succeed, after that a new initial crossover point is randomly selected in the first parent;

In this research, the probability of crossover is distributed according to a logistic normal distribution:

$$Pc(t+1) = 1 - \left(1 + \frac{1 - Pc(t)}{Pc(t)} \exp(-\gamma N(0,1))\right)^{-1}$$

The learning rate $\gamma$ controls the size of adaptation steps.

**Mutation operation:** we used Inversion Mutation, the inversion mutation [3] is randomly select a sub string_BNF, removes it from the string_BNF and inserts it in a randomly selected position. However, the sub string_BNF is inserted in reversed order. The mutation operation will create some new individuals that might not be produced by the reproduction and crossover operations. Generally, a lower probability of mutation can guarantee the convergence of the GA, but it may lead to a poor solution quality. On the other hand, a higher probability of mutation may lead to the phenomenon of a random walk for the GA. In this research, the probability of mutation is set to be $P_m=1-P_c$.

**Stop criterion**: the genetic algorithm repeatedly runs the reproduction, crossover, mutation, and replacement operations until it meets the stop criterion. The stop criterion is set to be 1000 generations, because this criterion can obtain satisfied solution.

## 4. Experimental Results

To illustrate the ability of the proposed method to determine the acceptable solutions SE, we present in this section the results of its application to two different examples of quantum systems, used as test examples. The first example is symmetric finite-barrier potential [4] whose potential is defined by:

$$V(x) = \begin{bmatrix} 0 & 0 < x < 1 \\ \infty & Otherwise \end{bmatrix}$$

With a=0, b=1, the wave function is the form:

$\Psi = e^{kx}$ Where $k^2 = \dfrac{2mE}{\hbar^2}$ and $E = \dfrac{\pi^2}{2}$

The second we calculate ground-state wave function of a particle of mass m in a harmonic potential with the potential $V(x) = \dfrac{1}{2}\omega^2 x^2$ Where $\omega = \sqrt{20} \times 10^2$

The wave function is :

$\Psi = e^{\dfrac{-x^2}{4\alpha^2}}$ Where $\alpha^2 = \dfrac{\hbar}{2m\omega}$

We used for the population size a number in [200,1000], the chromosome length is 50. The number of equidistant points T=100. The experiments were performed on a core duo 2400 Mhz, running in Debian Linux with c++ language. The algorithm found the same functions.

## 5. Conclusion

In this paper we have outlined and some initial results a evolutionary technique for solving the SE. This technique of hand the idea is to pose the problem of solving this equation in terms of problem research is to find the best wave function in the space of all possible wave functions. Test results of this technique are very satisfactory and encouraging further study.

**Khalid Jebari** received his B. Sc. degree in computer science in 1992. His M. Sc. degree in computer science, multimedia and telecommunications in 2008 and Ph. D. degree in computer sciences in 2013 both from Faculty of Science, Mohammed V Agdal University, Morocco. His research interests include genetic algorithms, particle swarm, fuzzy clustering, and E-learning.

**Mohammed Madiafi** received his B. Sc. degree in computer science in 2006. His M. Sc. degree in computer science, and mathematics in 2008 and Ph. D. degree in computer science in 2013 both from Faculty of Sciences, Ben M'sik Casablanca, Morocco. His research interests include Artificial Neural Network, fuzzy clustering, image processing, evolutionary computation.

**Abdelaziz Elmoujahid** received his B. Sc. degree in applied mathematics in 2000. His M. Sc. degree in computer science, multimedia and telecommunications in 2008 both from Faculty of Science, Mohammed V Agdal University, Morocco. He is a Ph. D. candidate in Faculty of Sciences Rabat. His research interests include genetic algorithms, fuzzy clustering, fuzzy learning, forecasting.